\documentclass{llncs}
\usepackage{graphicx}
\usepackage{xcolor}
\usepackage{url}
\usepackage{tabularx}
\makeatletter
\g@addto@macro{\UrlBreaks}{\UrlOrds}
\makeatother

\begin{document}

\title{On the relationship between Benchmarking, Standards and Certification in Robotics and AI}

\author{Alan F.T. Winfield \and Matthew Studley}
\institute{Bristol Robotics Lab, University of the West of England, Bristol, UK}


\maketitle              

\begin{abstract}
Benchmarking, standards and certification are closely related processes. Standards can provide normative requirements that robotics and AI systems may or may not conform to. Certification generally relies upon conformance with one or more standards as the key determinant of granting a certificate to operate. And benchmarks are sets of standardised tests against which robots and AI systems can be measured. Benchmarks therefore can be thought of as informal standards. In this paper we will develop these themes with examples from benchmarking, standards and certification, and argue that these three linked processes are not only useful but vital to the broader practice of Responsible Innovation.
\end{abstract}


\section{Introduction}

Benchmarking, standards and certification are closely related processes. Standards, especially those drafted and published by international standards bodies, can provide normative requirements that robotics and AI systems may or may not conform to. Certification, especially for safety-critical systems, generally relies upon conformance with one or more standards as the key determinant of granting a certificate to operate. And benchmarks are sets of standardised tests against which robots and AI systems can be measured. Benchmarks therefore can be thought of as informal standards, developed by robotics or AI communities without the formal drafting and approval structures of national and international standards bodies. 

In this paper we will develop and link these themes and argue that the three processes of benchmarking, standardisation and certification are not only useful but vital to the broader practice of Responsible Innovation. In sections 2 and 3 we define benchmarks and standards, and provide examples from robotics and AI. In section 4 we show how the three processes are linked within the frameworks of responsible innovation and ethical governance, and review current regulation and standards in robotics, including assistive robots, drones and AVs, then AI. Section 5 concludes with a discussion of the regulatory gaps identified by the work of this paper.

\section{Benchmarks}

The term benchmark was first used to refer to the practice of testing rifles by securely fixing them to a bench in order to compare the spread of several identical shots on a target. The benchmark was devised as a more reliable and repeatable method than employing a human marksman.

The Oxford English Dictionary defines a benchmark as ``something that can be measured and used as a standard that other things can be compared with". Perhaps more usefully JISC\footnote{JISC is the UK Joint Information Systems Committee, which provides networking, IT and high performance computing to colleges and universities.} defines benchmarks as ``reference points or measurements used for comparison, usually with the connotation that the benchmark is a ‘good’ standard against which comparison can be made" and benchmarking as ``a process of finding good practice and of learning from others". The JISC article on benchmarking also usefully makes the distinction between metric benchmarks and process benchmarks\footnote{https://www.jisc.ac.uk/guides/benchmarking/what-is-benchmarking}.


\begin{figure}[htbp]
\centering
\includegraphics[width=13.0cm]{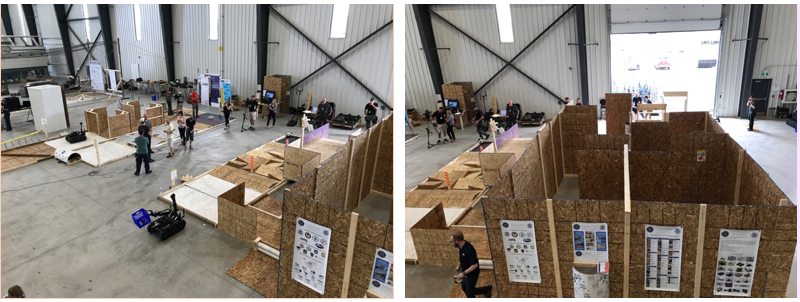}
\caption{Left: A basic suite of mobility and dexterity test methods. Right: The Access Buildings suite of test methods includes terrains, stoops, stairs, doors, and hallway Labyrinths all with embedded dexterity and mapping tasks. Credit: NIST}
\label{Fig:NIST}
\end{figure} 

For mobile robots the best known benchmarks are those developed and published by the US National Institute for Standards and Technology (NIST). Fig \ref{Fig:NIST} shows two test areas for mobile robot mobility and navigation. Notably individual tests are fabricated from wood. This significantly reduces the cost and difficulty of reproducing the tests. For an example of repeatable terrain tests for evaluating robot mobility see \cite{Jacoff2008}.

Robot competitions, especially those that are well established events, also provide benchmark tests. RoboCUP is perhaps the best known international robot competition\footnote{https://www.robocup.org/}. Held annually since 1997, the core events of RoboCUP are soccer matches between teams of robots. The event is split into leagues, based upon both the size of the robots and whether they are wheeled or legged. More recently RoboCUP has grown to encompass competitions for assisted living robots: RoboCUP@home, workplace robots: RoboCUP@work and rescue robots: RoboCUP@rescue\cite{Stone2023}.

Some robotics competitions have formally defined benchmarks, then used those benchmarks to determine the scores of competing teams. The EU RockIn competition, for instance, defined both task benchmarks (TBMs) and functionality benchmarks (FBMs). TBMs measure the performance of integrated robot systems for a specified task, while FBMs ``focus on the performance of single subsystems, defining a precise setup in which a single robot functionality can be evaluated''\cite{Amigoni2015}. The same approach was adopted in the EU euRathlon outdoor air, land and sea robot search and rescue competition\cite{Winf2016}, and subsequently the EU European Robotics League\footnote{https://eu-robotics.net/eurobotics/activities/european-robotics-league/}.

\section{Standards}

British standard BS 0 \textit{A standard for standards} formally defines a standard as a “document, established by consensus and approved by a recognized body, that provides, for common and repeated use, rules, guidelines or characteristics for activities or their results, aimed at the achievement of the optimum degree of order in a given context”, with the valuable note that ``Standards should be based on the consolidated results of science, technology and experience, and aimed at the promotion of optimum community benefits" \cite{bsi0}.

Technical standards fall into five broad categories, as summarised in Table \ref{tableDim} below.

\begin{table}[h]
\centering
    \begin{tabularx}{\textwidth}{|>{\raggedright\arraybackslash}X |>{\raggedright\arraybackslash}X| } 
    \hline
    Type & Meaning \\
    \hline\hline
    Standard Specification & An explicit set of requirements for an item, material, component, system or service.  \\ 
    \hline
    Standard Test Method & A definitive procedure that produces a test result. \\ 
    \hline
    Standard Practice or Procedure &  a set of instructions for performing operations or functions. \\
    \hline
    Standard Guidelines & General information or options that do not require a specific course of action. \\
    \hline
    Standard Definition & A formally established terminology \\ 
    \hline
    \end{tabularx}
    \caption{Five types of Technical Standard}
	\label{tableDim}
\end{table}

Typically the first 3 types in Table \ref{tableDim}: specifications, test methods and practices, are standards that can be conformed to or measured against. An example in robotics is ISO 13482:2014
\textit{Robots and robotic devices -- Safety requirements for personal care robots}, which sets out safety requirements \cite{iso13482}. Directly related is a recent standard setting out test methods ISO/TR 23482-1:2020
\textit{Robotics -- Application of ISO 13482 -- Part 1: Safety-related test methods} \cite{iso23482}. An example of a standard definition in robotics is the well known ISO 8373:2012 \textit{Robots and robotic devices - Vocabulary} \cite{iso8373}.

In recent years a new generation of explicitly ethical standards in robotics and AI are emerging \cite{winf19}. Almost certainly the first, British Standard BS 8611:2016 \textit{Guide to the ethical design and application of robots and robotic systems} is a good example of a standard setting out guidelines \cite{bsi16}.

The IEEE Standards Association Global Initiative on the ethics of Intelligent and Autonomous Systems\footnote{https://standards.ieee.org/industry-connections/ec/autonomous-systems/} has, to date, led to six published standards:
\begin{itemize}
    \item IEEE 7000-2021 \textit{Model Process for Addressing Ethical Concerns during System Design}
    \item IEEE 7001-2021 \textit{Transparency of Autonomous Systems}
    \item IEEE 7002-2022 \textit{Standard for Data Privacy Process}
    \item IEEE 7005-2021 \textit{Standard for Transparent Employer Data Governance}
    \item IEEE 7007-2021 \textit{Ontological Standard for Ethically Driven Robotics and Automation Systems}
    \item IEEE 7010-2020 \textit{Recommended Practice for Assessing the Impact of Autonomous and Intelligent Systems on Human Well-Being}
\end{itemize}

The majority of standards, including all of those mentioned above, are drafted and published by national and international standards bodies. These standards bodies have formalised processes for the structure and operation of working groups (which comprise expert volunteers) alongside rigorous processes of review and approval prior to publication. It is this level of rigour that substantially differentiates standards and benchmarks.

There is also a parallel tradition of Open Standards, which are openly accessible and usable by anyone. Notably most of the Internet protocols were drafted and published as open standards, more commonly known as Requests for Comment (RFCs)\footnote{See for instance https://www.rfc-editor.org/rfc/rfc822}. A recent open standard sets out a draft specification for an Ethical Black Box for social robots\cite{winf2022}.

\section{Linking benchmarks, standards and certification}

Certification is defined as ``the provision by an independent body of written assurance (a certificate) that the product, service or system in question meets specific requirements"\footnote{https://www.iso.org/certification.html}. When systems are safety critical then certification is often a legal requirement and it is typically a government agency, or regulator, that will either conduct or commission the process of certification. Thus regulation and certification are linked. Given that robots -- and the AIs embedded in them -- are safety critical systems, we shall here focus on certification as part of regulation.

\begin{figure}[htbp]
    \centering
    \includegraphics[width=12.5cm]{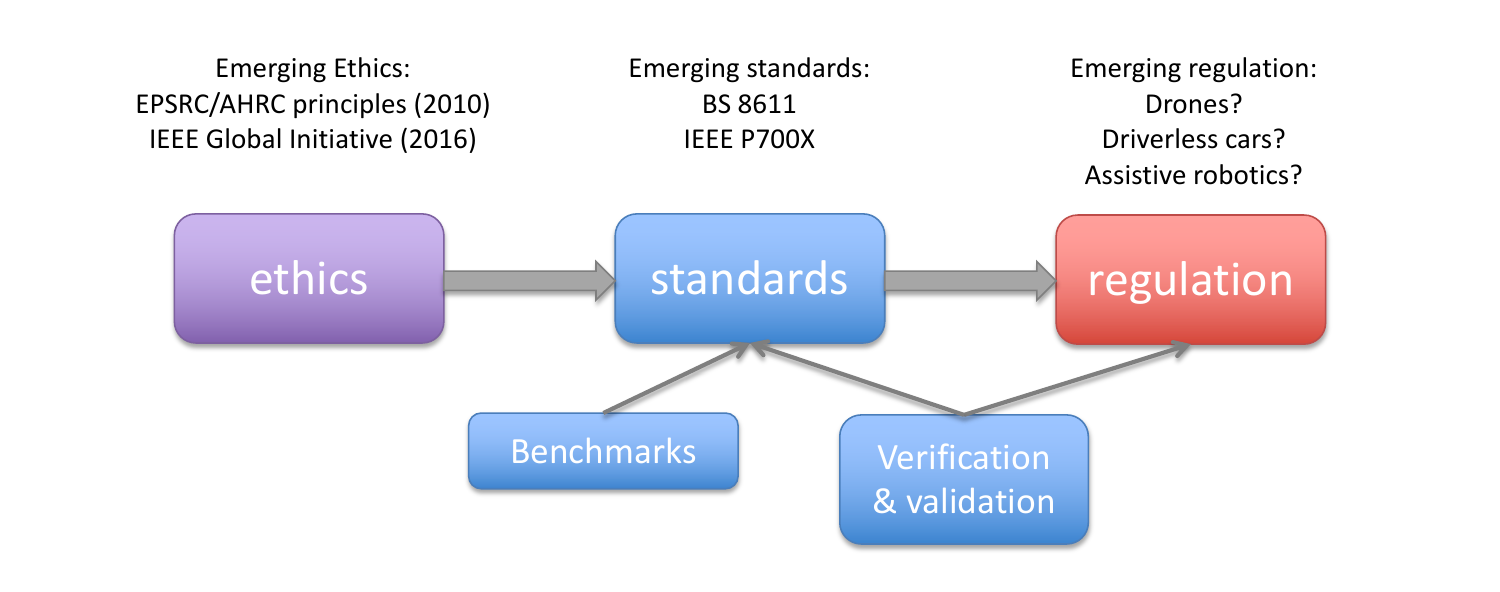}
    \caption{A schematic showing the link between ethics, benchmarks, standards and regulation, adapted from \cite{winf18}}
    \label{Fig:BenchStanFig}
\end{figure}

In a paper on ethical governance in robotics and AI \cite{winf18} we argue that ethics underpin standards, which in turn underpin regulation. This relationship is shown schematically in Figure \ref{Fig:BenchStanFig}. Responsible robotics and AI requires all of the processes in Figure \ref{Fig:BenchStanFig} \textit{et alia}; for a full account see \cite{winf18}. 

Also shown in Figure \ref{Fig:BenchStanFig} are benchmarks, which both compliment and underpin formal standards. And for completeness we show that the processes of verification and validation, which are essential to assuring the safety of real-world systems, underpin both standards and regulation.

In order to understand the relationship between standards and regulation, we first need to appreciate that all standards are voluntary. British standard BS 0 makes this very clear: ``British Standards are voluntary in that there is no obligation to apply them or comply with them, except in those few cases where their application is directly demanded by regulatory instruments" \cite{bsi0}. Usefully adding: standards ``are tools devised for the convenience of those who wish to use them. In certain circumstances the actions of third parties might have the effect of making the application of a standard a commercial necessity, e.g. in a contract."

Even though standards are voluntary compliance with regulation is not. Regulators rely on standards to provide specifications against which systems or processes can be measured and certified. As an example take electrical installations in buildings. British Standard BS 7671:2018+A2:2022 sets out \textit{Requirements for electrical Installations}. As the rubric for this standard states BS 7671 ``is the most widely accepted way to demonstrate compliance with the applicable legislation, and so if you are an electrical professional, you need to ensure that you are working to the ... latest version of this standard" \cite{bsi7671}.

\subsection{Regulation and related standards in Robotics}

Consider first industrial robots. In the UK the Health and Safety Executive (HSE) is the regulator for health, safety and welfare in the workplace. The HSE was created following an act of parliament: the Health and Safety at Work, etc. Act 1974. The HSE is also responsible for investigating industrial accidents. The relevant standards here are EN ISO 10218-1:2011
\textit{Robots and robotic devices -- Safety requirements for industrial robots -- Part 1: Robots} \cite{iso10218pt1}, and \textit{Part 2: Robot systems and integration} \cite{iso10218pt2}. Since both hospitals and care homes are workplaces then personal care robots will arguably also fall within the regulatory orbit of the HSE. Such robots require the application of standard ISO 13482:2014 \textit{Robots and robotic devices -- Safety requirements for personal care robots} \cite{iso13482} and the more recent test methods set out in ISO/TR 23482-1:2020 \textit{Robotics -- Application of ISO 13482 -- Part 1: Safety-related test methods} \cite{iso23482}. It is important to note that when the same personal care robots specified in ISO 13482 are used in private homes, they are not subject to any regulatory oversight.

Consider now Unmanned Air Vehicles, or drones, which are regulated by the same bodes that regulate civil aviation. In the UK for instance the Civil Aviation Authority (CAA) regulates both commercial unmanned aircraft and drone operations, and recreational unmanned aircraft. In November 2019 it became illegal to fly or operate without first registering and taking a theory test. Drones must also be labeled with their operator ID. The relevant standards, which cover both commercial and consumer drones, are ISO 21384 \textit{Unmanned aircraft systems Part 2: UAS Components} and \textit{Part 3: Operational Procedures} \cite{iso21384}.

Another topical area is Autonomous Vehicles (AVs). At present AVs do not have a dedicated regulatory body but are regulated by existing transport agencies, for instance in the USA, this is the National Highway Traffic Safety Agency (NHTSA). Notably, serious AV accidents in the USA are investigated by the National Transport Safety Board (NTSB), the same body that investigates air accidents. In the UK, AVs fall under Highways England (monitored by the Office of Rail and Road). Regulation and certification of AVs remains a work in progress. In the US policy is more advanced at state level than federal, with Michigan, California and Arizona leading \cite{Savrin2022}. Europe, according to a report in \cite{Juliussen2022}, is behind the US and China in AV technology and testing yet is leading in both standards and regulation. Key recently published AV standards include UL 4600 \textit{Standard for Safety for the Evaluation of Autonomous Products} (2020)\footnote{A second edition was published in March 2022} and ISO 21448:2022 \textit{Road vehicles — Safety of the intended functionality}. ISO Technical Committee 22 on Road Vehicles published a Report on prospective standardisation for AVs in January 2021\footnote{https://www.iso.org/home.isoDocumentsDownload.do?t=yU9HibyZYru2MLwGlSJAUwaO9lAqZXaoaw-3FAKrR-ooVMtpoemScl2jrFIG8BLy}. 

\subsection{Regulation and related standards in AI}

At the time of writing AI regulation is a work on progress. AI regulation that is  closest to becoming law is the EU's AI Act. First proposed in April 2021, the AI Act's approach is risk based with the aim of making sure that ``AI systems used in the EU are safe, transparent, traceable, non-discriminatory and environmentally friendly"\footnote{https://www.europarl.europa.eu/news/en/headlines/society/20230601STO93804/eu-ai-act-first-regulation-on-artificial-intelligence}. The Act was adopted by the European Parliament in June 2023, and talks with member states on the final form of the law are in progress and expected to conclude before the end of 2023. For a summary of the
significance and scope of the AI Act see \cite{Edwards2022}. A recent article in the New York Times reported that ``the United States remains far behind Europe" and ``is only at the beginning of what is likely to be a long and difficult path toward the creation of AI rules" \cite{Kang2023}.

Fortunately AI Standards development has been underway at pace since 2018. The ISO/IEC joint technical committee 1/SC 42 on Artificial Intelligence has published 20 AI standards since 2018, and a further 30 are in development\footnote{see https://www.iso.org/committee/6794475/x/catalogue/}. Published standards notably include ISO/IEC 22989:2022
\textit{Information technology -- Artificial intelligence -- Artificial intelligence concepts and terminology} \cite{iso22989}, ISO/IEC 38507:2022 \textit{Information technology -- Governance of IT -- Governance implications of the use of artificial intelligence by organizations} \cite{iso38507} and ISO/IEC 25059:2023 \textit{Software engineering -- Systems and software Quality Requirements and Evaluation (SQuaRE) -- Quality model for AI systems} \cite{iso25059}. When taken together with the six IEEE 7000 series standards listed in section 3 above, a substantial corpus of AI standards can be called upon by regulators and certification agencies. 

\section{Discussion and conclusions}

In this paper we have outlined definitions for benchmarks, standards and certification, alongside example benchmarks and standards in robotics. We have also shown how benchmarks, standards and certification are linked processes, underpinned by ethics. As argued in \cite{winf19} all benchmarks and standards embody the ethical principle that shared good practice benefits all. Certification and regulation importantly provide both assurance and public confidence that technical systems are safe.

We have identified a gap in the regulatory landscape: the use of robots in private homes. These robots could be assisted living or personal care robots, smart robot toys, or utility robots such as vacuum cleaning or lawn-mowing robots. It follows that if accidents at home in which robots are implicated, there are no accident investigation agencies with responsibility for discovering what happened, how the robot failed, and what measures are needed to prevent future similar accidents. In a recent paper we argued that the likelihood of accidents involving social robots\footnote{robots that work with or in close proximity to humans} is much higher than for industrial robots, and that there is a an urgent need for both thorough processes of social robot accident investigation and agencies responsible for such investigations \cite{Winfield2021}.

There is no doubt that benchmarks, standards and certification are essential instruments of responsible robotics and AI.

\section*{Acknowledgments}
The work of this paper has been conducted within EPSRC project RoboTIPS, grant reference EP/S005099/1 \textit{RoboTIPS: Developing Responsible Robots for the Digital Economy} 

\bibliographystyle{ieeetr}
\bibliography{main}

\begin{thebibliography}{10}

\bibitem{Jacoff2008}
A.~Jacoff, A.~Downs, A.~Virts, and E.~Messina, ``Stepfield pallets: Repeatable
  terrain for evaluating robot mobility,'' in {\em Proceedings of the 2008
  Performance Metrics for Intelligent Systems (PerMIS) Workshop, Gaithersburg,
  MD}, 2008-12-31 2008.

\bibitem{Stone2023}
P.~Stone, ``Will robots triumph over world cup winners by 2050?,'' {\em IEEE
  Spectrum}, June 2023.

\bibitem{Amigoni2015}
F.~Amigoni, E.~Bastianelli, J.~Berghofer, A.~Bonarini, G.~Fontana,
  N.~Hochgeschwender, L.~Iocchi, G.~Kraetzschmar, P.~Lima, M.~Matteucci,
  P.~Miraldo, D.~Nardi, and V.~Schiaffonati, ``Competitions for benchmarking:
  Task and functionality scoring complete performance assessment,'' {\em IEEE
  Robotics \& Automation Magazine}, vol.~22, no.~3, pp.~53--61, 2015.

\bibitem{Winf2016}
A.~F.~T. Winfield, M.~P. Franco, B.~Brueggemann, A.~Castro, M.~C. Limon,
  G.~Ferri, F.~Ferreira, X.~Liu, Y.~Petillot, J.~Roning, F.~Schneider,
  E.~Stengler, D.~Sosa, and A.~Viguria, ``eu{R}athlon 2015: A multi-domain
  multi-robot {G}rand {C}hallenge for {S}earch and {R}escue {R}obots,'' in {\em
  Towards Autonomous Robotic Systems} (L.~Alboul, D.~Damian, and J.~M. Aitken,
  eds.), (Cham), pp.~351--363, Springer International Publishing, 2016.

\bibitem{bsi0}
BSI, {\em BS0:2011 A Standard for Standards – Principles of standardization}.
\newblock British Standards Institute, 2011.

\bibitem{iso13482}
ISO, {\em ISO 13482:2014 Robots and robotic devices — Safety requirements for
  personal care robots}.
\newblock International Standards Organisation, 2014.

\bibitem{iso23482}
ISO, {\em ISO/TR 23482-1:2020 Robotics — Application of ISO 13482 — Part 1:
  Safety-related test methods}.
\newblock International Standards Organisation, 2020.

\bibitem{iso8373}
ISO, {\em ISO 8373:2021 Robotics — Vocabulary}.
\newblock International Standards Organisation, 2021.

\bibitem{winf19}
A.~Winfield, ``Ethical standards in robotics and {AI},'' {\em Nature
  Electronics}, vol.~2, no.~2, pp.~46--48, 2019.

\bibitem{bsi16}
BSI, {\em BS8611:2016 Robots and robotic devices, Guide to the ethical design
  and application of robots and robotic systems}.
\newblock British Standards Institute, 2016.

\bibitem{winf2022}
A.~Winfield, A.~van Maris, P.~Salvini, and M.~Jirotka, ``An {E}thical {B}lack
  {B}ox for {S}ocial {R}obots: a draft open standard,'' in {\em Proc. Intl.
  Conf. on Robot Ethics and Standards (ICRES 2022)}, 18-19 July 2022.

\bibitem{winf18}
A.~F. Winfield and M.~Jirotka, ``Ethical governance is essential to building
  trust in robotics and artificial intelligence systems, phil,'' {\em Trans. R.
  Soc. A}, vol.~376, no.~20180085, 2018.

\bibitem{bsi7671}
BSI, {\em BS7671:2018+A2:2022 Requirements for electrical Installations. IET
  Wiring Regulations}.
\newblock British Standards Institute, 2022.

\bibitem{iso10218pt1}
ISO, {\em ISO 10218-1:2011 Robots and robotic devices -- Safety requirements
  for industrial robots -- Part 1: Robots}.
\newblock International Standards Organisation, 2018.

\bibitem{iso10218pt2}
ISO, {\em ISO 10218-2:2011 Robots and robotic devices — Safety requirements
  for industrial robots -- Part 1: Robot Systems and Integration}.
\newblock International Standards Organisation, 2018.

\bibitem{iso21384}
ISO/DIS, {\em ISO 21384-2:2021 Unmanned aircraft systems — Part 2: UAS
  components}.
\newblock ISO, 2021.

\bibitem{Savrin2022}
D.~Savrin and M.~Fanelli, ``State—not federal—policy guides us autonomous
  driving,'' {\em Automotive World}, September 2022.

\bibitem{Juliussen2022}
E.~Juliussen, ``Autonomous vehicles: How is {E}urope doing?,'' {\em EE Times
  Europe}, November 2022.

\bibitem{Edwards2022}
L.~Edwards, ``The {EU AI} {A}ct: a summary of its significance and scope,''
  {\em The Ada Lovelace Institute: expert explainer}, April 2022.

\bibitem{Kang2023}
C.~Kang, ``In {U.S.}, {R}egulating {A.I.} is in its ‘early days’,'' {\em
  The New York Times}, July 2023.

\bibitem{iso22989}
ISO/IEC, {\em ISO/IEC 22989:2022 Information technology -- Artificial
  intelligence -- Artificial intelligence concepts and terminology}.
\newblock ISO/IEC JTC 1/SC 42, 2022.

\bibitem{iso38507}
ISO/IEC, {\em ISO/IEC 38507:2022 Information technology -- Governance of IT --
  Governance implications of the use of artificial intelligence by
  organizations}.
\newblock ISO/IEC JTC 1/SC 42, 2022.

\bibitem{iso25059}
ISO/IEC, {\em ISO/IEC 25059:2023 Software engineering -- Systems and software
  Quality Requirements and Evaluation (SQuaRE) -- Quality model for AI
  systems}.
\newblock ISO/IEC JTC 1/SC 42, 2023.

\bibitem{Winfield2021}
A.~F.~T. Winfield, K.~Winkle, H.~Webb, U.~Lyngs, M.~Jirotka, and C.~Macrae,
  {\em Robot Accident Investigation: A Case Study in Responsible Robotics},
  pp.~165--187.
\newblock Cham: Springer International Publishing, 2021.

\end{thebibliography}

\end{document}